\pdfoutput=1
\documentclass[11pt]{article}
\usepackage{eacl2017}
\usepackage{times}
\usepackage{url}
\usepackage{latexsym}
\usepackage{amssymb}
\usepackage{latexsym}
\usepackage{subfigure} 
\usepackage{multirow}
\usepackage{url} 

\usepackage{breakurl}
\usepackage[utf8]{inputenc}
\usepackage{graphicx}
\usepackage{array}
\usepackage{amsmath}
\usepackage[font=singlespacing]{caption}
\usepackage{fancyhdr}
\usepackage{amssymb}
\usepackage{nccmath}
\usepackage{color}
 \usepackage[normalem]{ulem}
 \usepackage{lipsum}
 \usepackage{graphics}
 \usepackage{tikz-dependency}
 \usepackage{enumitem}
\usepackage{tikz-qtree}
\usepackage{tikz}
 \usepackage{pgfplots}
  \usepackage{pgfplotstable}
  \usepackage{relsize}
 
\setitemize{noitemsep,topsep=10pt,parsep=0pt,partopsep=0pt}
\setenumerate{noitemsep,topsep=10pt,parsep=0pt,partopsep=0pt}

\newcommand{\ignore}[1]{}
\newenvironment{itemizesquish}{\begin{list}{\setcounter{enumi}{0}\labelitemi}{\setlength{\itemsep}{-0.25em}\setlength{\labelwidth}{0.5em}\setlength{\leftmargin}{\labelwidth}\addtolength{\leftmargin}{\labelsep}}}{\end{list}}

\eaclfinalcopy 
\def\eaclpaperid{446} 

\newcommand{\ask}[1]{\textcolor{orange}{{\textbf{[#1 --\textsc{ask}]}}}}
\newcommand{\lpk}[1]{\textcolor{green}{{\textbf{[#1 --\textsc{lpk}]}}}}
\newcommand{\cjd}[1]{\textcolor{cyan}{{\textbf{[#1 --\textsc{cjd}]}}}}
\newcommand{\nascomment}[1]{\textcolor{blue}{{\textbf{[#1 --\textsc{nas}]}}}}
\newcommand{\miguelcomment}[1]{\textcolor{red}{{\textbf{[#1 --\textsc{miguel}]}}}}
\newcommand{\grahamcomment}[1]{\textcolor{pink}{\textbf{[#1 --\textsc{graham}]}}}

 \renewcommand{\ask}[1]{}
 \renewcommand{\lpk}[1]{}
 \renewcommand{\cjd}[1]{}
\renewcommand{\nascomment}[1]{}
 \renewcommand{\miguelcomment}[1]{}
 \renewcommand{\grahamcomment}[1]{}

\title{What Do Recurrent Neural Network Grammars Learn About Syntax?}
\author{Adhiguna Kuncoro$^{\spadesuit}$ ~ Miguel Ballesteros$^{\diamondsuit}$  ~ Lingpeng Kong$^{\spadesuit}$  \\ 
\textbf{Chris Dyer}$^{\spadesuit\clubsuit}$ ~ \textbf{Graham Neubig}$^{\spadesuit}$ ~~ \textbf{Noah A. Smith}$^{\heartsuit}$ \\
$^{\spadesuit}$School of Computer Science, Carnegie Mellon University, Pittsburgh, PA, USA \\
$^{\diamondsuit}$IBM T.J. Watson Research Center, Yorktown Heights, NY, USA \\
$^{\clubsuit}$DeepMind, London, UK\\
$^{\heartsuit}$Computer Science \& Engineering, University of Washington, Seattle, WA, USA\\
{\small \tt \{akuncoro,lingpenk,gneubig\}@cs.cmu.edu}\\ {\small \tt miguel.ballesteros@ibm.com, cdyer@google.com, nasmith@cs.washington.edu}
}

\date{}

\begin{document}
\maketitle
\begin{abstract}
Recurrent neural network grammars (RNNG) are a recently proposed probabilistic generative modeling family for natural language. They show state-of-the-art language modeling and parsing performance. We investigate what information they learn, from a linguistic perspective, through various ablations to the model and the data, and by augmenting the model with an attention mechanism (GA-RNNG) to enable closer inspection. We find that explicit modeling of composition is crucial for achieving the best performance. Through the attention mechanism, we find that headedness plays a central role in phrasal representation (with the model's latent attention largely agreeing with predictions made by hand-crafted head rules, albeit with some important differences). By training grammars without nonterminal labels, we find that phrasal representations depend minimally on nonterminals, providing support for the endocentricity hypothesis. 
\end{abstract}

\section{Introduction}

In this paper, we focus on a recently proposed class of probability distributions,
recurrent neural network grammars (RNNGs; \nocite{rnng} Dyer et al., 2016), designed to model syntactic derivations of sentences.  We focus on RNNGs as generative probabilistic models over trees, as summarized in \S\ref{sec:rnng}.

Fitting a probabilistic model to data has often been understood as a way to test or confirm some aspect of a theory.  We talk about a model's assumptions and sometimes explore its parameters or posteriors over its latent variables in order to gain understanding of what it ``discovers'' from the data.  In some sense,  such models can be thought of as mini-scientists.

Neural networks, including RNNGs, are capable of representing larger classes of hypotheses than traditional probabilistic models, giving them more freedom to explore.  Unfortunately, they tend to be bad mini-scientists, because their parameters are difficult for human scientists to interpret.

RNNGs are striking because they obtain state-of-the-art parsing and language modeling performance.  Their relative lack of independence assumptions, while still incorporating a degree of linguistically-motivated prior knowledge, affords the model considerable freedom to derive its own insights about syntax. If they are mini-scientists, the discoveries they make should be of particular interest as propositions about syntax (at least for the particular genre and dialect of the data).

This paper manipulates the inductive bias of RNNGs to test linguistic hypotheses.\footnote{RNNGs have less inductive bias relative to traditional unlexicalized probabilistic context-free grammars, but more than models that parse by transducing word sequences to linearized parse trees represented as strings \cite{vinyals:2015}.  Inductive bias is necessary for learning \cite{mitchell-80}; we believe the important question is not ``how little can a model get away with?'' but rather the benefit of different forms of inductive bias as data vary.}
We begin with an ablation study to discover the importance of the composition function in \S\ref{sec:composition_analysis}. Based on the findings, we augment the RNNG composition function with a novel \textbf{gated attention} mechanism (leading to the GA-RNNG) to incorporate more interpretability into the model in \S\ref{sec:ga_rnng}. Using the GA-RNNG, we proceed by investigating the role that individual heads play in phrasal representation (\S\ref{sec:headedness}) and the role that nonterminal category labels play (\S\ref{sec:bracketing}). Our key findings are that lexical heads play an important role in representing most phrase types (although compositions of multiple salient heads are not infrequent, especially for conjunctions) and that nonterminal labels provide little additional information. As a by-product of our investigation, a variant of the RNNG without ensembling achieved the best reported supervised phrase-structure parsing (93.6 $F_1$; English PTB) and, through conversion, dependency parsing (95.8 UAS, 94.6 LAS; PTB SD). The code and pretrained models to replicate our results are publicly available\footnote{\url{https://github.com/clab/rnng/tree/master/interpreting-rnng}}.

\section{Recurrent Neural Network Grammars}\label{sec:rnng}

An RNNG defines a \emph{joint} probability distribution over string terminals and phrase-structure nonterminals.\footnote{\newcite{rnng} also defined a conditional version of the RNNG that can be used only for parsing; here we focus on the generative version since it is more flexible and (rather surprisingly) even learns better estimates of $p(\boldsymbol{y} \mid \boldsymbol{x})$.} Formally, the RNNG is defined by a triple $\langle N,\Sigma,\Theta \rangle$, where $N$ denotes the set of nonterminal symbols (NP, VP, etc.), $\Sigma$ the set of all terminal symbols (we assume that $N \cap \Sigma = \emptyset$), and $\Theta$ the set of all model parameters. Unlike previous works that rely on hand-crafted rules to compose more fine-grained phrase representations \cite{collins_97,klein_03}, the RNNG implicitly parameterizes the information passed through compositions of phrases (in $\Theta$ and the neural network architecture), hence weakening the strong independence assumptions in classical probabilistic context-free grammars. 

The RNNG is based on an abstract state machine like those used in transition-based parsing, with its algorithmic state consisting of a stack of partially completed constituents, a buffer of already-generated terminal symbols, and a list of past actions. To generate a sentence $\boldsymbol{x}$ and its phrase-structure tree $\boldsymbol{y}$, the RNNG samples a sequence of actions to construct $\boldsymbol{y}$ top-down. Given $\boldsymbol{y}$, there is one such sequence (easily identified), which we call the oracle, $\boldsymbol{a} = \langle a_1,\ldots,a_n \rangle$ used during supervised training.

The RNNG uses three different actions:
\begin{itemizesquish}
\item $\textsc{nt}(\mathrm{X})$, where $\mathrm{X} \in N$, introduces an open nonterminal symbol onto the stack, e.g., ``(NP'';
\item $\textsc{gen}(\mathrm{x})$, where $\mathrm{x} \in \Sigma$, generates a terminal symbol and places it on the stack and buffer; and
\item $\textsc{reduce}$ indicates a constituent is now complete.  The elements of the stack that comprise the current constituent (going back to the last open nonterminal) are popped, a \textbf{composition function} is executed, yielding a composed representation that is pushed onto the stack.
\end{itemizesquish}

\begin{figure}
\centering
\vspace{-.2cm}\includegraphics[scale=.3]{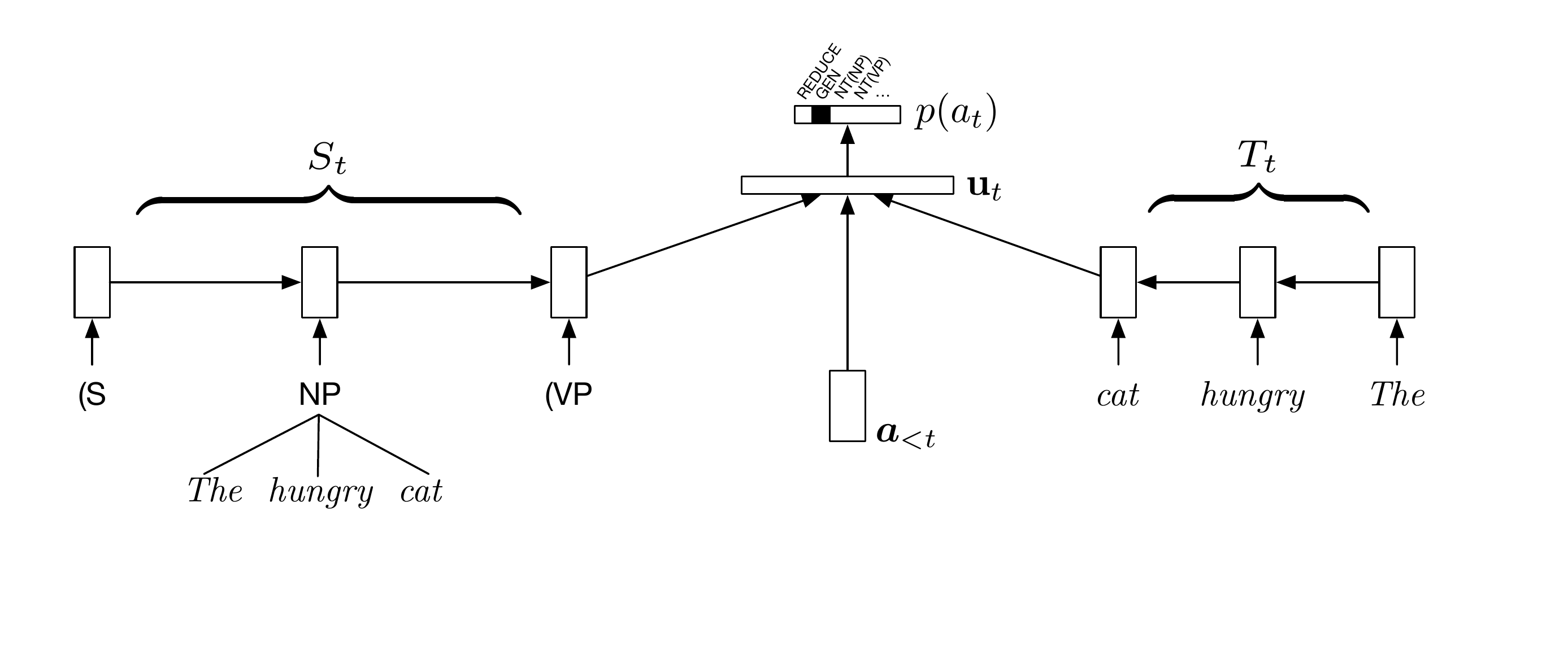}
\vspace{-.7cm}\caption{The RNNG consists of a stack, buffer of generated words, and list of past actions that lead to the current configuration. Each component is embedded with LSTMs, and the parser state summary $\mathbf{u}_t$ is used as top-layer features to predict a softmax over all feasible actions. This figure is due to \newcite{rnng}.
\label{fig:genstate}}
\end{figure}

At each timestep, the model encodes the stack, buffer, and past actions, with a separate LSTM \cite{hochreiter_97} for each component as features to define a distribution over the next action to take (conditioned on the full algorithmic state).  
The overall architecture is illustrated in Figure \ref{fig:genstate}; examples of full action sequences can be found in \newcite{rnng}.

\begin{figure}[h]
\centering
\includegraphics[scale=.45]{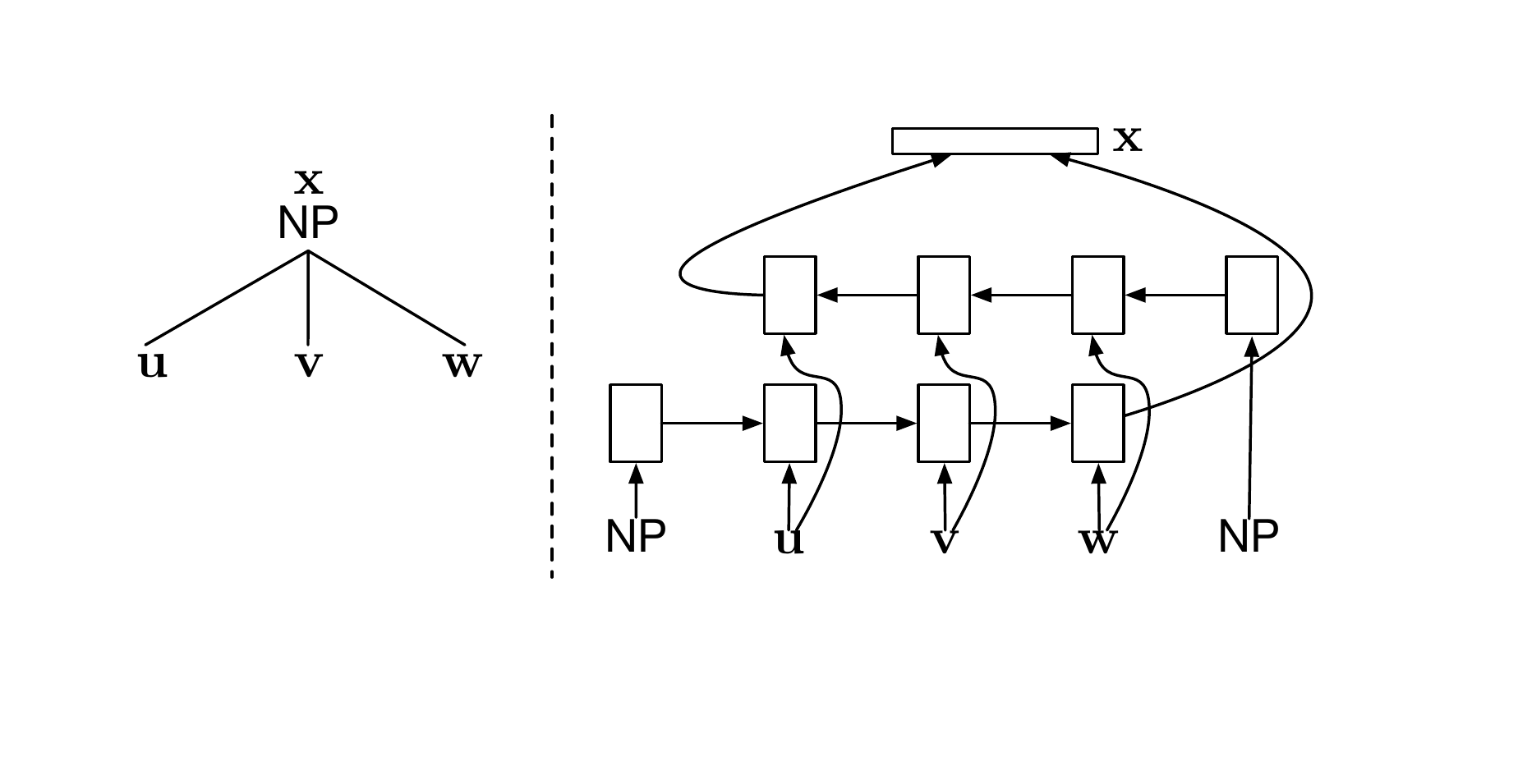}
\vspace{-1.3cm}
\caption{RNNG composition function on each $\textsc{reduce}$ operation; the network on the right models the structure on the left \cite{rnng}.}
\label{fig:composition}
\end{figure}

A key element of the RNNG is the composition function, which reduces a completed constituent into a single element on the stack.  This function computes a vector representation of the new constituent; it also uses an LSTM (here a bidirectional one).  This composition function, which we consider in greater depth in \S\ref{sec:composition_analysis}, is illustrated in Fig.~\ref{fig:composition}.

Since the RNNG is a generative model, it attempts to maximize $p(\boldsymbol{x}, \boldsymbol{y})$, the \emph{joint} distribution of strings and trees, defined as
\begin{align*}
p(\boldsymbol{x}, \boldsymbol{y}) = p(\boldsymbol{a}) = \prod_{t=1}^{n} p(a_t \mid a_1,\ldots,a_{t-1}).
\end{align*}
In other words, $p(\boldsymbol{x}, \boldsymbol{y})$ is defined as a product of local probabilities, conditioned on all past actions. The joint probability estimate $p(\boldsymbol{x}, \boldsymbol{y})$ can be used for both phrase-structure parsing (finding $\arg\max_{\boldsymbol{y}} p(\boldsymbol{y} \mid \boldsymbol{x})$)  and language modeling (finding $p(\boldsymbol{x})$ by marginalizing over the set of possible parses for $\boldsymbol{x}$).  Both inference problems can be solved using an importance sampling procedure.\footnote{Importance sampling works by using a proposal distribution $q(\boldsymbol{y} \mid \boldsymbol{x})$ that is easy to sample from. In \newcite{rnng} and this paper, the proposal distribution is the discriminative variant of the RNNG; see \newcite{rnng}.} We report all RNNG performance based on the corrigendum to \newcite{rnng}.

\ignore{\miguelcomment{I think sect. 2 is too long, we have already published a paper with that info. If you want to do it like this, you should add some discussion about the things that you want to study/change in this paper... research questions,etc.}
\nascomment{disagree.  section 1 is way too long -- coming back to that later.  section 2 needs to make the paper self-contained since our contributions rely on digging into the model details}\miguelcomment{True. I "ignored" both comments}}

\section{Composition is Key}\label{sec:composition_analysis}

Given the same data, under both the discriminative and generative settings
 RNNGs were found to parse with significantly higher accuracy than (respectively) the models of \newcite{vinyals:2015} and \newcite{choe:2016} that represent $\boldsymbol{y}$ as a ``linearized'' sequence of symbols and parentheses without explicitly capturing the tree structure, or even constraining the $\boldsymbol{y}$ to be a well-formed tree (see Table~\ref{tab:vinyalscomp}). \newcite{vinyals:2015} directly predict the sequence of nonterminals, ``shifts'' (which consume a terminal symbol), and parentheses from left to right, conditional on the input terminal sequence $\boldsymbol{x}$, while \newcite{choe:2016} used a sequential LSTM language model on the same linearized trees to create a generative variant of the \newcite{vinyals:2015} model. The generative model is used to re-rank parse candidates.

\begin{table}[h]
      \centering
       \begin{tabular}{l|r}
		\textbf{Model} & $F_1$ \\
		\hline
		\newcite{vinyals:2015} -- PTB only & 88.3 \\
		Discriminative RNNG & \textbf{91.2} \\ \hline \hline
        \newcite{choe:2016} -- PTB only & 92.6 \\
        Generative RNNG & \textbf{93.3} 
		\end{tabular}
        \caption{Phrase-structure parsing performance on PTB \S 23. All results are reported using single-model performance and without any additional data.}
        \label{tab:vinyalscomp}
      \end{table}

The results in Table~\ref{tab:vinyalscomp} suggest that the RNNG's explicit composition function (Fig.~\ref{fig:composition}), which \newcite{vinyals:2015} and \newcite{choe:2016} must learn implicitly, plays a crucial role in the RNNG's generalization success. Beyond this, Choe and Charniak's generative variant of Vinyals et al.~(2015) is another instance where generative models trained on the PTB outperform discriminative models.

\subsection{Ablated RNNGs}
On close inspection, it is clear that the RNNG's three data structures---stack, buffer, and action history---are redundant.
For example, the action history and buffer contents completely determine the structure of the stack at every timestep. Every generated word goes onto the stack, too; and some past words will be composed into larger structures, but through the composition function, they are all still ``available'' to the network that predicts the next action. Similarly, the past actions are redundant with the stack.  Despite this redundancy, only the stack incorporates the composition function. Since each of the ablated models is sufficient to encode all necessary partial tree information, the primary difference is that ablations with the stack use explicit composition, to which we can therefore attribute most of the performance difference.

We conjecture that the stack---the component that makes use of the composition function---is critical to the RNNG's performance, and that the buffer and action history are not.  In transition-based parsers built on expert-crafted features, the most recent words and actions are useful if they are salient, although neural representation learners can automatically learn what information should be salient.

To test this conjecture, we train \textbf{ablated RNNGs} that lack each of the three data structures (action history, buffer, stack), as well as one that lacks both the action history and buffer.\footnote{Note that the ablated RNNG without a stack is quite similar to \newcite{vinyals:2015}, who encoded a (partial)
 phrase-structure tree as a sequence of open and close parentheses, terminals, and nonterminal symbols; our action history is quite close to this, with each \textsc{nt}(X) capturing a left parenthesis and X nonterminal, and each \textsc{reduce} capturing a right parenthesis.}
If our conjecture is correct, performance should degrade most without the stack, and the stack alone should perform competitively.

\textbf{Experimental settings.} We perform our experiments on the English PTB corpus, with \S02--21 for training, \S24 for validation, and \S23 for test; no additional data were used for training. We follow the same hyperparameters as the generative model proposed in \newcite{rnng}.\footnote{The model is trained using stochastic gradient descent, with a learning rate of 0.1 and a per-epoch decay of 0.08. All experiments with the generative RNNG used 100 tree samples for each sentence, obtained by sampling from the local softmax distribution of the discriminative RNNG.} The generative model did not use any pretrained word embeddings or POS tags; a discriminative variant of the standard RNNG was used to obtain tree samples for the generative model. All further experiments use the same settings and hyperparameters unless otherwise noted.

\textbf{Experimental results.} 
We trained each ablation from scratch, and compared these models on three tasks:  English phrase-structure parsing (labeled $F_1$), Table~\ref{tab:parsing}; dependency parsing, Table~\ref{tab:dep}, by converting parse output to Stanford dependencies \cite{demarneffe-06} using the tool by \newcite{kong_14}; and language modeling, Table~\ref{tab:ppl}.  The last row of each table reports the performance of a novel variant of the (stack-only) RNNG with attention, to be presented in \S\ref{sec:ga_rnng}.

\begin{table}[!htb]
      \centering
       \begin{tabular}{l|r}
		\textbf{Model} & $F_1$ \\
		\hline
		\newcite{vinyals:2015}$^{\dagger}$ & 92.1 \\
        \newcite{choe:2016} & 92.6 \\
        \newcite{choe:2016}$^{\dagger}$ & \textbf{93.8} \\
		Baseline RNNG  & 93.3 \\
		\hline
		\hline
		Ablated RNNG (no history)  & 93.2 \\
		Ablated RNNG (no buffer) &  93.3 \\
		Ablated RNNG (no stack)  & 92.5 \\
		Stack-only RNNG &  \textbf{93.6} \\
		\hline
		\hline
		GA-RNNG & 93.5 
		\end{tabular}
        \caption{Phrase-structure parsing performance on PTB \S 23. $^{\dagger}$ indicates systems that use additional unparsed data (semisupervised). The GA-RNNG results will be discussed in \S\ref{sec:ga_rnng}.}
        \label{tab:parsing}
      \end{table}
        
 \begin{table}[!htb]
 \scalebox{0.85}{
       \begin{tabular}{l|r|r}
		\textbf{Model} & \textbf{UAS} & \textbf{LAS} \\
		\hline
		\newcite{kiperwasser}&  93.9 & 91.9 \\
		\newcite{globally_normalized} & 94.6 & 92.8 \\
        \newcite{dozat:2016} & 95.4 & 93.8 \\ 
        \newcite{choe:2016}$^{\dagger}$ & \textbf{95.9} & 94.1 \\
		Baseline RNNG & 95.6 & 94.4 \\
		\hline
		\hline
		Ablated RNNG (no history)  & 95.4 & 94.2 \\
		Ablated RNNG (no buffer) &  95.6 & 94.4 \\
		Ablated RNNG (no stack)  & 95.1 & 93.8 \\
		Stack-only RNNG &  \textbf{95.8} & \textbf{94.6} \\
		\hline
		\hline
		GA-RNNG & 95.7 & 94.5
		\end{tabular}
        }
        \caption{Dependency parsing performance on PTB \S23 with Stanford Dependencies \cite{stanford_dependencies}. $^{\dagger}$ indicates systems that use additional unparsed data (semisupervised).} 
        \label{tab:dep}
\end{table}

\textbf{Discussion.}
The RNNG with only a stack is the strongest of the ablations, and it even outperforms the ``full'' RNNG with all three data structures.  Ablating the stack gives the worst among the new results.  This strongly supports the importance of the composition function:  a proper \textsc{reduce} operation that transforms a constituent's parts and nonterminal label into a single explicit (vector) representation is helpful to performance.

It is noteworthy that the stack alone is stronger than the original RNNG, which---in principle---can learn to disregard the buffer and action history.  Since the stack maintains syntactically ``recent'' information near its top, we conjecture that the learner is overfitting to spurious predictors in the buffer and action history that explain the training data but do not generalize well.

A similar performance degradation is seen in language modeling (Table~\ref{tab:ppl}):  the stack-only RNNG achieves the best performance, and ablating the stack is most harmful.  Indeed, modeling syntax without explicit composition (the stack-ablated RNNG) provides little benefit over a sequential LSTM language model.


\begin{table}[h]
\begin{center}
\scalebox{0.85}{
\begin{tabular}{l|c}
\textbf{Model} & \textbf{Test ppl.} (PTB)  \\
\hline
IKN 5-gram & 169.3 \\
LSTM LM & 113.4   \\
RNNG &  105.2 \\
\hline
\hline
Ablated RNNG (no history) & 105.7 \\
Ablated RNNG (no buffer) & 106.1 \\
Ablated RNNG (no stack) & 113.1 \\
Stack-only RNNG & \textbf{101.2} \\
\hline
\hline
GA-RNNG & \textbf{100.9}
\end{tabular}
}
\end{center}
\caption{Language modeling:  perplexity. IKN refers to Kneser-Ney 5-gram LM.
\label{tab:ppl}}
\end{table}%

We remark that the stack-only results are the best published PTB results for both phrase-structure and dependency parsing among supervised models.
 
\section{Gated Attention RNNG}\label{sec:ga_rnng}


Having established that the composition function is key to RNNG performance 
 (\S\ref{sec:composition_analysis}), we now seek to understand the nature of the composed phrasal representations that are learned. Like most neural networks, interpreting the composition function's behavior is challenging. Fortunately, linguistic theories offer a number of hypotheses about the nature of representations of phrases that can provide a conceptual scaffolding to understand them.

\subsection{Linguistic Hypotheses}
We consider two theories about phrasal representation. The first is that phrasal representations are strongly determined by a privileged lexical head. Augmenting grammars with lexical head information has a long history in parsing, starting with the models of \newcite{collins_97}, and theories of syntax such as the ``bare phrase structure'' hypothesis of the Minimalist Program~\cite{chomsky:1993} posit that phrases are represented purely by single lexical heads. Proposals for multiple headed phrases (to deal with tricky cases like conjunction) likewise exist \cite{jackendoff:1977,keenan:1987}. Do the phrasal representations learned by RNNGs depend on individual lexical heads or multiple heads? Or do the representations combine all children without any salient head?
 
Related to the question about the role of heads in phrasal representation is the question of whether phrase-internal material wholly determines the representation of a phrase (an endocentric representation) or whether nonterminal relabeling of a constitutent introduces new information (exocentric representations). To illustrate the contrast, an endocentric representation is representing a noun phrase with a noun category, whereas $\text{S} \rightarrow \text{NP}\ \text{VP}$ exocentrically introduces a new syntactic category that is neither NP nor VP \cite{chomsky:1970}.

\subsection{Gated Attention Composition}
To investigate what the stack-only RNNG learns about headedness (and later endocentricity), we propose a variant of the composition function that makes use of an explicit attention mechanism \cite{bahdanau_15} and a sigmoid gate with multiplicative interactions, henceforth called \textbf{GA-RNNG}. 

At every $\textsc{reduce}$ operation, the GA-RNNG assigns an ``attention weight'' to each of its children (between 0 and 1 such that the total weight off all children sums to 1), and the parent phrase is represented by the combination of a sum of each child's representation scaled by its attention weight and its nonterminal type. Our weighted sum is more expressive than traditional head rules, however, because it allows attention to be divided among multiple constituents. Head rules, conversely, are analogous to giving all attention to one constituent, the one containing the lexical head.

We now formally define the GA-RNNG's composition function.
Recall that $\mathbf{u}_t$ is the concatenation of the vector representations of the RNNG's data structures, used to assign probabilities to each of the actions available at timestep $t$ (see Fig.~\ref{fig:genstate}, the layer before the softmax at the top). For simplicity, we drop the timestep index here. Let $\mathbf{o}_{\mathit{nt}}$ denote the vector embedding (learned) of the nonterminal being constructed, for the purpose of computing attention weights. 

Now let $\mathbf{c}_1, \mathbf{c}_2,\ldots$ denote the sequence of vector embeddings for the constituents of the new phrase.  The length of these vectors is defined by the dimensionality of the bidirectional LSTM used in the original composition function (Fig.~\ref{fig:composition}). We use semicolon (;) to denote vector concatenation operations.

The attention vector is given by:
\begin{align}
\mathbf{a} = \mathrm{softmax}\left( \left[ \mathbf{c}_1 \ \mathbf{c}_2 \ \cdots \right]^{\top} \mathbf{V} \left[ \mathbf{u}; \mathbf{o}_{\mathit{nt}} \right]\right)
\end{align}
Note that the length of $\mathbf{a}$ is the same as the number of constituents, and that this vector sums to one due to the softmax.  It divides a single unit of attention among the constituents.

Next, note that the \emph{constituent source vector} $\mathbf{m} = [\mathbf{c}_1 ; \mathbf{c}_2; \cdots ] \mathbf{a}$ is a convex combination of the child-constituents, weighted by attention.  We will combine this with another embedding of the nonterminal denoted as $\mathbf{t}_{\mathit{nt}}$ (separate from $\mathbf{o}_{\mathit{nt}}$) using a sigmoid gating mechanism:
\begin{align}
\mathbf{g} & = \sigma \left( \mathbf{W}_1 \mathbf{t}_{\mathit{nt}} + \mathbf{W}_2 \mathbf{m} + \mathbf{b} \right)
\end{align}
Note that the value of the gate is bounded between $\left[0, 1 \right]$ in each dimension.

The new phrase's final representation uses element-wise multiplication ($\odot$) with respect to both $\mathbf{t}_{\mathit{nt}}$ and $\mathbf{m}$, a process reminiscent of the LSTM ``forget'' gate:
\begin{align}
\mathbf{c} &= \mathbf{g} \odot \mathbf{t}_{\mathit{nt}} + \left( 1 -  \mathbf{g}\right) \odot \mathbf{m}.
\end{align}

The intuition is that the composed representation should incorporate both nonterminal information and information about the constituents (through weighted sum and attention mechanism). The gate $\mathbf{g}$ modulates the interaction between them to account for varying importance between the two in different contexts.  
\ignore{\ask{add figure here. Or not, if the equations and intuitions are clear enough; I tried making it as clear as possible since some reviewers may not be familiar with attention (mostly used in MT). Any thoughts?}. \nascomment{I don't think we need a figure.  I tried to make it even cleaner; please check carefully.  I got rid of some symbols.}}


\textbf{Experimental results.} We include this model's performance in Tables \ref{tab:parsing}--\ref{tab:ppl} (last row in all tables). It is clear that the model outperforms the baseline RNNG with all three structures present and achieves competitive performance with the strongest, stack-only, RNNG variant.

\section{Headedness in Phrases}\label{sec:headedness}
We now exploit the attention mechanism to probe what the RNNG learns about headedness on the WSJ \S 23 test set (unseen before by the model).
\subsection{The Heads that GA-RNNG Learns}\label{sec:heads_GA_RNNG}

\nascomment{need to figure out -- perplexity or entropy?}\cjd{strongly in favor of ppl}

The attention weight vectors tell us which constituents are most important to a phrase's vector representation in the stack.  Here, we inspect the attention vectors to investigate whether the model learns to center its attention around a single, or by extension a few, salient elements, which would confirm the presence of headedness in GA-RNNG.

First, we consider several major nonterminal categories, and estimate the average \emph{perplexity} of the attention vectors. The average perplexity can be interpreted as the average number of ``choices'' for each nonterminal category; this value is only computed for the cases where the number of components in the composed phrase is at least two (otherwise the attention weight would be trivially 1). The minimum possible value for the perplexity is 1, indicating a full attention weight around one component and zero everywhere else. 

\begin{figure}[!h]
\centering
\begin{tikzpicture}
\scalebox{0.85}{
    \begin{axis}[
            ybar,
            symbolic x coords={ADJP, VP, NP, PP, QP, SBAR},
            xtick=data,
        ]
        \addplot table[x=NT,y=H(a)]{\mydata};
        \addplot table[x=NT,y=H(u)]{\mydata};
    \end{axis}
    }
\end{tikzpicture}
\caption{Average perplexity of the learned attention vectors on the test set (blue), as opposed to the average perplexity of the uniform distribution (red), computed for each major phrase type. \nascomment{perplexity, or entropy?} }
\label{fig:entropy}
\end{figure}
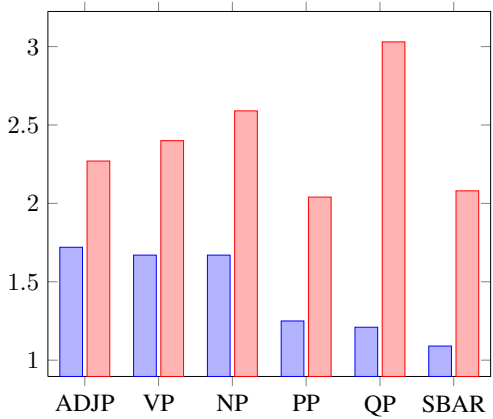

Figure~\ref{fig:entropy} (in blue) shows much less than 2 average ``choices'' across nonterminal categories, which also holds true for all other categories not shown. For comparison we also report the average perplexity of the uniform distribution for the same nonterminal categories (Fig.~\ref{fig:entropy} in red); this represents the highest entropy cases where there is no headedness at all by assigning the same attention weight to each constituent (e.g. attention weights of 0.25 each for phrases with four constituents). It is clear that the learned attention weights have much lower perplexity than the uniform distribution baseline, indicating that the learned attention weights are quite peaked around certain components. This implies that phrases' vectors tend to resemble the vector of one salient constituent, but not exclusively, as the perplexity for most categories is still not close to one. 

Next, we consider the how attention is distributed for the major nonterminal categories in Table~\ref{tab:attention}, where the first five rows of each category represent compositions with highest entropy, and the next five rows are qualitatively analyzed.  The high-entropy cases where the attention is most divided represent more complex phrases with conjunctions or more than one plausible head.\ask{Make sure this is not controversial}

\textbf{NPs.} In most simple noun phrases (representative samples in rows 6--7 of Table~\ref{tab:attention}), the model pays the most attention to the \emph{rightmost noun} and assigns near-zero attention on determiners and possessive determiners, while also paying nontrivial attention weights to the adjectives. This finding matches existing head rules and our intuition that nouns head noun phrases, and that adjectives are more important than determiners. 

We analyze the case where the noun phrase contains a conjunction in the last three rows of Table \ref{tab:attention}. The syntax of conjunction is a long-standing source of controversy in syntactic analysis~\cite[\emph{inter alia}]{johannessen:1998}. Our model suggests that several representational strategies are used, when coordinating single nouns, both the first noun (8) and the last noun (9) may be selected. However, in the case of conjunctions of multiple noun \emph{phrases} (as opposed to multiple single-word nouns), the model consistently picks the conjunction as the head. All of these representational strategies have been argued for individually on linguistic grounds, and since we see all of them present, RNNGs face the same confusion that linguists do.

\textbf{VPs.} The attention weights on simple verb phrases (e.g., ``VP $\to$ V NP'', 9) are peaked around the noun phrase instead of the verb. This implies that the verb phrase would look most similar to the noun under it and contradicts existing head rules that unanimously put the verb as the head of the verb phrase. 
Another interesting finding is that the model pays attention to \emph{polarity} information, where negations are almost always assigned non-trivial attention weights.\footnote{Cf.~\newcite{li_16}, where sequential LSTMs discover polarity information in sentiment analysis, although perhaps more surprising as polarity information is less intuitively central to syntax and language modeling.} Furthermore, we find that the model attends to the conjunction terminal in conjunctions of verb phrases (e.g., ``VP $\to$ VP and VP'', 10), reinforcing the similar finding for conjunction of noun phrases.

\textbf{PPs.} In almost all cases, the model attends to the preposition terminal instead of the noun phrases or complete clauses under it, regardless of the type of preposition. Even when the prepositional phrase is only used to make a connection between two noun phrases (e.g., ``PP $\to$ NP after NP'', 10), the prepositional connector is still considered the most salient element. This is less consistent with the Collins and Stanford head rules, where prepositions are assigned a lower priority when composing PPs, although more consistent with the Johansson head rule \cite{Johansson:07}.\ask{To my understanding there is some contention on what the head of a preposition phrase should be. Some might say that the preposition symbol is not important and we should instead pay attention to whatever is under it, but some others say that the preposition symbol is indeed the head. I'm not sure how to formulate this, we should probably cite some previous works that talk and discuss about this.} 

\pgfplotstableread[row sep=\\,col sep=&]{
    NT & H(a) & H(u)   \\
    ADJP  & 1.72 & 2.27  \\
    VP     & 1.67 & 2.40  \\
    NP    & 1.67 & 2.59   \\
    PP     & 1.25 & 2.04 \\
    QP   & 1.21 & 3.03  \\
    SBAR    & 1.09 & 2.08 \\
    }\mydata

\begin{table*}[!ht]
\centering
  \resizebox{2\columnwidth}{!}{
\begin{tabular}{r|l|l|l|}
\multicolumn{1}{c}{} &\multicolumn{1}{c}{\textbf{Noun phrases}}                                   & \multicolumn{1}{c}{\textbf{Verb phrases}}           & \multicolumn{1}{c}{\textbf{Prepositional phrases}}          \\ \cline{2-4}
\scriptsize 1 & Canadian (0.09) \textbf{Auto (0.31)} Workers (0.2) union (0.22) president (0.18)  & \textbf{buying (0.31)} and (0.25) selling (0.21) NP (0.23) & ADVP (0.14) \textbf{on (0.72)} NP (0.14) \\
\scriptsize 2 & no (0.29) major (0.05) \textbf{Eurobond (0.32)} or (0.01) foreign (0.01) bond (0.1) offerings (0.22) & ADVP (0.27) \textbf{show (0.29)} PRT (0.23) PP (0.21) & ADVP (0.05) \textbf{for (0.54)} NP (0.40) \\
\scriptsize 3 & Saatchi (0.12) client (0.14) Philips (0.21) Lighting (0.24) \textbf{Co. (0.29)}& \textbf{pleaded (0.48)} ADJP (0.23) PP (0.15) PP (0.08) PP (0.06) & ADVP (0.02) \textbf{because (0.73)} of (0.18) NP (0.07) \\
\scriptsize 4 & nonperforming (0.18) commercial (0.23) \textbf{real (0.25)} estate (0.1) \textbf{assets (0.25)} & \textbf{received (0.33)} PP (0.18) NP (0.32) PP (0.17) & such (0.31) \textbf{as (0.65)} NP (0.04) \\
\scriptsize 5 &  the (0.1) Jamaica (0.1) Tourist (0.03) Board (0.17) ad (0.20) \textbf{account (0.40)} & cut (0.27) \textbf{NP (0.37)} PP (0.22) PP (0.14) & from (0.39) \textbf{NP (0.49)} PP (0.12) \\ \cline{2-4}\cline{2-4}
\scriptsize 6 &  the (0.0) final (0.18) \textbf{hour (0.81)}                       & \textbf{to (0.99)} VP (0.01)              & \textbf{of (0.97)} NP (0.03)             \\
\scriptsize 7 &  their (0.0) first (0.23) \textbf{test (0.77)}                     & \textbf{were (0.77)} n't (0.22) VP (0.01) & \textbf{in (0.93)} NP (0.07)             \\
\scriptsize 8 &\textbf{Apple (0.62)} , (0.02) Compaq (0.1) and (0.01) IBM (0.25) & did (0.39) \textbf{n't (0.60)} VP (0.01)  & \textbf{by (0.96)} S (0.04)              \\
\scriptsize 9 &  both (0.02) stocks (0.03) and (0.06) \textbf{futures (0.88)}      & handle (0.09) \textbf{NP (0.91)}          & \textbf{at (0.99)} NP (0.01)             \\
\scriptsize 10 &  NP (0.01) , (0.0) \textbf{and (0.98)} NP (0.01)                   & VP (0.15) \textbf{and (0.83)} VP 0.02)    & NP (0.1) \textbf{after (0.83)} NP (0.06) \\ \cline{2-4}
\end{tabular}
  }
\caption{Attention weight vectors for some representative samples for NPs, VPs, and PPs.}
\label{tab:attention}
\end{table*}

\subsection{Comparison to Existing Head Rules}

To better measure the overlap between the attention vectors and existing head rules, we converted the trees in PTB \S23 into a dependency representation using the attention weights. In this case, the attention weight functions as a ``dynamic'' head rule, where all other constituents within the same composed phrase are considered to modify the constituent with the highest attention weight, repeated recursively. The head of the composed representation for ``S'' at the top of the tree is attached to a special root symbol and functions as the head of the sentence. \miguelcomment{Also, see my comments about Stanford, Collins and Johansson's above... it seems that we are probably more similar to Johansson's. I sent the conversion tool months ago. Not sure if there is time to evaluate there.}

We measure the overlap between the resulting tree and conversion results of the same trees using the \newcite{collins_97} and Stanford dependencies \cite{demarneffe-06} head rules. Results are evaluated using the standard evaluation script (excluding punctuation) in terms of UAS, since the attention weights do not provide labels.

\textbf{Results.} The model has a higher overlap with the conversion using Collins head rules (49.8 UAS) rather than the Stanford head rules (40.4 UAS). We attribute this large gap to the fact that the Stanford head rules incorporate more semantic considerations, while the RNNG is a purely syntactic model. In general, the attention-based tree output has a high error rate ($\approx$ 90\%) when the dependent is a verb, since the constituent with the highest attention weight in a verb phrase is often the noun phrase instead of the verb, as discussed above. The conversion accuracy is better for nouns ($\approx$ 50\% error), and much better for determiners (30\%) and particles (6\%) with respect to the Collins head rules.

\textbf{Discussion.}  GA-RNNG has the power to infer head rules, and to a large extent, it does.  It follows some conventions that are established in one or more previous head rule sets (e.g., prepositions head prepositional phrases, nouns head noun phrases), but attends more to a verb phrase's nominal constituents than the verb.  It is important to note that this is not the by-product of learning a specific model for parsing; the training objective is \emph{joint} likelihood, which is not a proxy loss for parsing performance.  These decisions were selected because they make the data maximally likely (though admittedly only locally maximally likely).  We leave deeper consideration of this noun-centered verb phrase hypothesis to future work.

\section{The Role of Nonterminal Labels}\label{sec:bracketing}

Emboldened by our finding that GA-RNNGs learn a notion of headedness, we next explore whether heads are sufficient to create representations of phrases (in line with an endocentric theory of phrasal representation) or whether extra nonterminal information is necessary.  If the endocentric hypothesis is true (that is, the representation of a phrase is built from \emph{within} depending on its components but independent of explicit category labels), then the nonterminal types should be easily inferred given the endocentrically-composed representation, and that ablating the nonterminal information would not make much difference in performance. Specifically, we train a GA-RNNG on unlabeled trees (only bracketings without nonterminal types), denoted U-GA-RNNG.  

This idea has been explored in research on methods for learning syntax with less complete annotation \cite{pereira-92}.   A key finding from \newcite{klein_02} was that, given bracketing structure, simple dimensionality reduction techniques could reveal conventional nonterminal categories with high accuracy; \newcite{petrov_2006} also showed that latent variables can be used to recover fine-grained nonterminal categories.  We therefore expect that the vector embeddings of the constituents that the U-GA-RNNG correctly recovers (on test data) will capture categories similar to those in the Penn Treebank. 

\textbf{Experiments.} Using the same hyperparameters and the PTB dataset, we first consider \emph{unlabeled} $F_1$ parsing accuracy.  On test data (with the usual split), the GA-RNNG achieves 94.2\%, while the U-GA-RNNG achieves 93.5\%. This result suggests that nonterminal category labels add a relatively small amount of information compared to purely endocentric representations.

\begin{figure}[h]
\centering
\includegraphics[width=0.37\textwidth,angle=90]{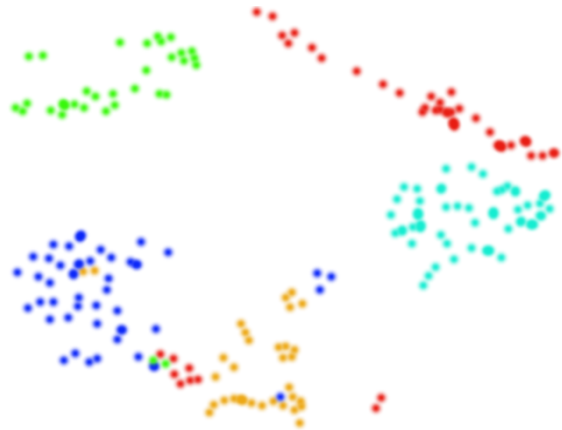}
\caption{t-SNE on composed vectors when training without nonterminal categories. Vectors in dark blue are VPs, red are SBARs, yellow are PPs, light blue are NPs, and green are Ss. 
\label{fig:nt-cluster}}
\end{figure}

\textbf{Visualization.} If endocentricity is largely sufficient to account for the behavior of phrases, where do our robust intuitions for syntactic category types come from? We use t-SNE \cite{maaten_08} to visualize composed phrase vectors from the U-GA-RNNG model applied to the unseen test data.
Fig. \ref{fig:nt-cluster} shows that the U-GA-RNNG tends to recover nonterminal categories as encoded in the PTB, even when trained without them.\footnote{We see a similar result for the non-ablated GA-RNNG model, not shown for brevity.} These results suggest nonterminal types can be inferred from the purely endocentric compositional process to a certain extent, and that the phrase clusters found by the U-GA-RNNG largely overlap with nonterminal categories.  

\begin{figure}[h]
\includegraphics[scale=0.25]{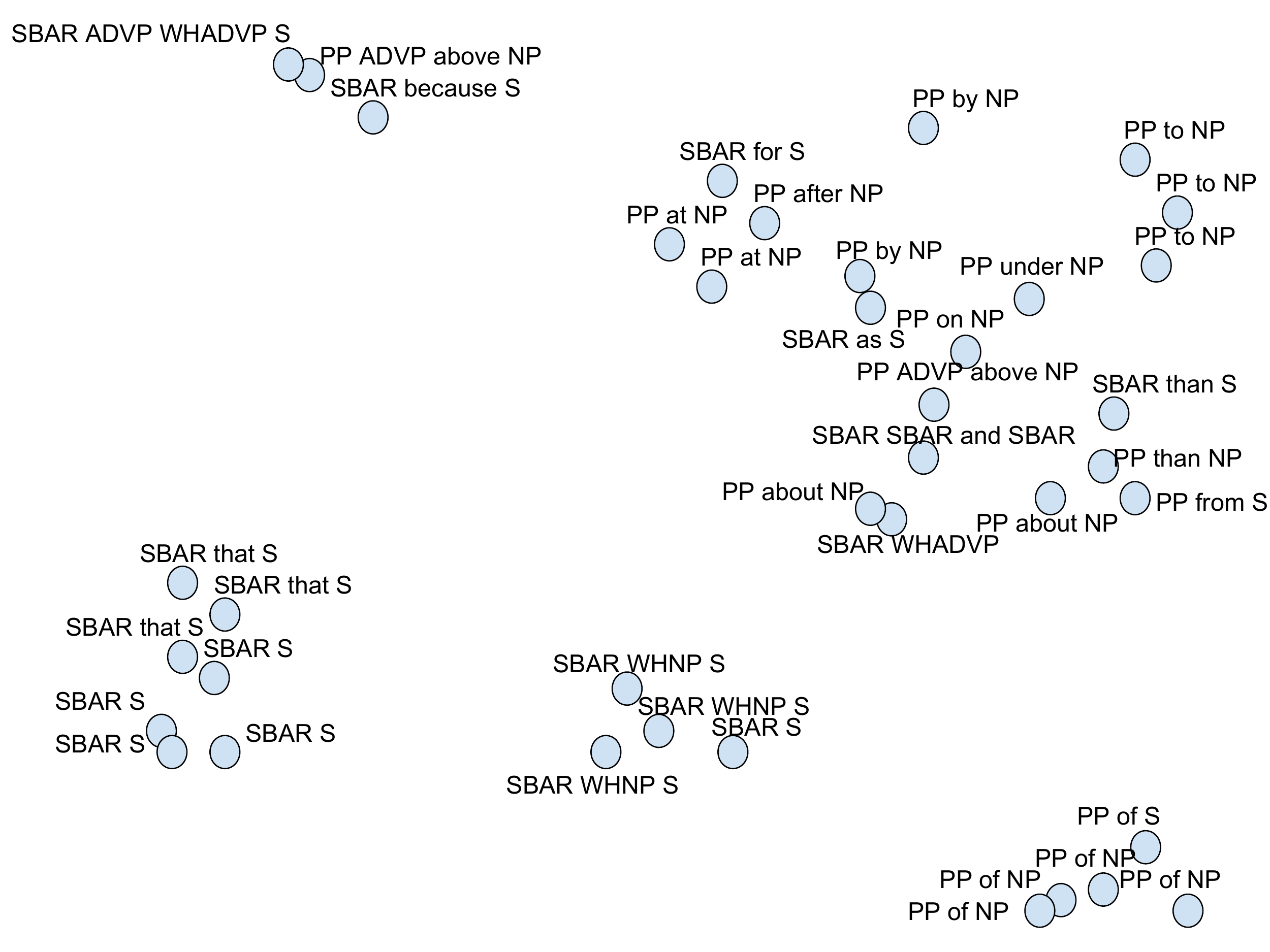}
\caption{Sample of PP and SBAR phrase representations.
\label{fig:pp-sbar}}
\end{figure}

\textbf{Analysis of PP and SBAR.} Figure \ref{fig:nt-cluster} indicates a certain degree of overlap between SBAR (red) and PP (yellow). As both categories are interesting from the linguistic perspective and quite similar, we visualize the learned phrase vectors of 40 randomly selected SBARs and PPs from the test set (using U-GA-RNNG), illustrated in Figure \ref{fig:pp-sbar}. First, we can see that phrase representations for PPs and SBARs depend less on the nonterminal categories\footnote{Recall that U-GA-RNNG is trained without access to the nonterminal labels; training the model with nonterminal information would likely change the findings.} and more on the connector. For instance, the model learns to cluster phrases that start with words that can be either prepositions or complementizers (e.g., \emph{for}, \emph{at}, \emph{to}, \emph{under}, \emph{by}), regardless of whether the true nonterminal labels are PPs or SBARs. This suggests that SBARs that start with ``prepositional'' words are similar to PPs from the model's perspective.

Second, the model learns to disregard the word \emph{that}, as ``$\text{SBAR} \rightarrow \textit{that}\ \text{S}$'' and ``$\text{SBAR} \rightarrow \text{S}$'' are close together. This finding is intuitive, as complementizer \emph{that} is often optional~\cite{jaeger:2010}, unlike prepositional words that might describe relative time and location. Third, certain categories of PPs and SBARs form their own separate clusters, such as those that involve the words \emph{because} and \emph{of}. We attribute these distinctions to the fact that these words convey different meanings than many prepositional words; the word \emph{of} indicates possession while \emph{because} indicates cause-and-effect relationship. These examples show that, to a certain extent, the GA-RNNG is able to learn non-trivial semantic information, even when trained on a fairly small amount of syntactic data.

\section{Related Work}\label{sec:related_works}
The problem of understanding neural network models in NLP has been previously studied for sequential RNNs \cite{karpathy_15,li_16}. 
\newcite{shi_16} showed that sequence-to-sequence neural translation models capture a certain degree of syntactic knowledge of the source language, such as voice (active or passive) and tense information, as a by-product of the translation objective. Our experiment on the importance of composition function was motivated by \newcite{vinyals:2015} and \newcite{wiseman_16}, who achieved competitive parsing accuracy without explicit composition. In another work, \newcite{li:15} investigated the importance of recursive tree structures (as opposed to linear recurrent models) in four different tasks, including sentiment and semantic relation classification. Their findings suggest that recursive tree structures are beneficial for tasks that require identifying long-range relations, such as semantic relationship classification, with no conclusive advantage for sentiment classification and discourse parsing. Through the stack-only ablation we demonstrate that the RNNG composition function is crucial to obtaining state-of-the-art parsing performance.

Extensive prior work on phrase-structure parsing typically employs the probabilistic context-free grammar formalism, with lexicalized \cite{collins_97} and nonterminal \cite{johnson_98,klein_03} augmentations. The conjecture that fine-grained nonterminal rules and labels can be discovered given weaker bracketing structures was based on several studies \cite{chiang_02,klein_02,petrov_2006}. 

In a similar work, \newcite{sangati:09} proposed entropy minimization and greedy familiarity maximization techniques to obtain lexical heads from labeled phrase-structure trees in an unsupervised manner. In contrast, we used neural attention to obtain the ``head rules'' in the GA-RNNG; the whole model is trained end-to-end to maximize the log probability of the correct action given the history. Unlike prior work, GA-RNNG allows the attention weight to be divided among phrase constituents, essentially propagating (weighted) headedness information from multiple components. 

\nascomment{be sure to discuss chiang and bikel; petrov and other latent variable PCFG papers}
\lpk{one thing make slav's paper interesting is that he find basically that some of the nonterminal categories need to be more fine-grained than others' (table 1 in http://www.petrovi.de/data/acl06.pdf). It would be interesting if we see something like this in our finding.}

\section{Conclusion}\label{sec:conclusion}
We probe what recurrent neural network grammars learn about syntax, through ablation scenarios and a novel variant with a gated attention mechanism on the composition function. The composition function, a key differentiator between the RNNG and other neural models of syntax, is crucial for good performance. Using the attention vectors we discover that the model is learning something similar to heads, although the attention vectors are not completely peaked around a single component. We show some cases where the attention vector is divided and measure the relationship with existing head rules. RNNGs without access to nonterminal information during training are used to support the hypothesis that phrasal representations are largely endocentric, and a visualization of representations shows that traditional nonterminal categories fall out naturally from the composed phrase vectors. This confirms previous conjectures that bracketing annotation does most of the work of syntax, with nonterminal categories easily discoverable given bracketing.

\section*{Acknowledgments}\begin{small}
This work was sponsored in part by the Defense Advanced Research Projects Agency (DARPA)
Information Innovation Office (I2O) under the Low Resource Languages for Emergent Incidents (LORELEI) program issued by DARPA/I2O under Contract No.~HR0011-15-C-0114;
it was also supported in part by Contract No.~W911NF-15-1-0543 with DARPA and the Army Research Office (ARO). Approved for public release, distribution unlimited. The views expressed are those of the authors and do not reflect the official policy or position of the Department of Defense or the U.S.~Government.

\end{small}

\bibliography{eacl2017}
\bibliographystyle{eacl2017}

\appendix

\end{document}